\documentclass{article} 

\usepackage{preprint}

\usepackage{cite}
\usepackage{amsmath,amssymb,amsfonts}
\usepackage{algorithmic}
\usepackage{graphicx}
\usepackage{textcomp}
\usepackage{xcolor}

\usepackage{url}
\usepackage{subcaption}
\usepackage{tabularx,booktabs}
\def\BibTeX{{\rm B\kern-.05em{\sc i\kern-.025em b}\kern-.08em
		T\kern-.1667em\lower.7ex\hbox{E}\kern-.125emX}}

\newcommand{\lowerromannumeral}[1]{\romannumeral#1\relax}

\usepackage{tikz,xcolor,hyperref}

\definecolor{lime}{HTML}{A6CE39}
\DeclareRobustCommand{\orcidicon}{
	\begin{tikzpicture}
	\draw[lime, fill=lime] (0,0)
	circle [radius=0.16]
	node[white] {{\fontfamily{qag}\selectfont \tiny ID}};
	\draw[white, fill=white] (-0.0625,0.095)
	circle [radius=0.007];
	\end{tikzpicture}
	\hspace{-2mm}
}
\foreach \x in {A, ..., Z}{\expandafter\xdef\csname orcid\x\endcsname{\noexpand\href{https://orcid.org/\csname orcidauthor\x\endcsname}
			{\noexpand\orcidicon}}
}

\title{Automatic Speech Recognition for Sanskrit with Transfer Learning}

\usepackage{authblk}

\author[1\thanks{\tt{sadhukhanbidit@gmail.com}}]{Bidit Sadhukhan}
\author[1\thanks{\tt{punyeshwarananda@gm.rkmvu.ac.in}}]{Swami Punyeshwarananda}

\affil[1]{Department of Computer Science, 
		
	 Ramakrishna Mission Vivekananda Educational and Research Institute,
	 
	 Belur Math, Howrah, India
 }

\begin{document}


\maketitle

\begin{abstract}
Sanskrit, one of humanity's most ancient languages, has a vast collection of books and manuscripts on diverse topics that have been accumulated over millennia. However, its digital content  (audio and text), which is vital for the training of AI systems, is profoundly limited. Furthermore, its intricate linguistics make it hard to develop robust NLP tools for wider accessibility. Given these  constraints, we have developed an automatic speech recognition model for Sanskrit by employing transfer learning mechanism on OpenAI's Whisper model.  After carefully optimising the hyper-parameters, we obtained promising results with our transfer-learned model achieving a word error rate of 15.42\% on Vāksañcayaḥ dataset.
An online demo of our model is made available for the use of public and to evaluate its performance firsthand thereby paving the way for improved accessibility and technological support for Sanskrit learning in the modern era.
\end{abstract}
%

\section{Introduction}
Sanskrit is one of the oldest languages in the world with a rich literary tradition.
A vast collection of ancient texts such as the Vedas, Upanishads, epic poetry (e.g., Ramayana and Mahabharata)\cite{awasthi1965}, and other scientific texts on topics such as mathematics, medicine and astronomy\cite{singh2006} were all composed in Sanskrit. The language served as a unifying thread for Indian culture and knowledge for centuries. Its influence permeates not only on modern Indian languages such as Hindi, Bengali, Kannada, and Marathi, but also on southeast Asian languages like Thai, Burmese and Javanese~\cite{malik1995,stargardt2003}.

The modern era poses numerous challenges to the preservation of Sanskrit,
as its usage in both spoken and written forms is limited in today's world~\cite{dalmia2013}. The paucity of resources in digital form also hinders its accessibility and learning for a broader audience. 
Recent efforts to revitalise the language, particularly its spoken form, through educational initiatives, contemporary literature, and advancements in technology, are ongoing~\cite{bhardwaj2018}. 

Traditional methods of learning and focusing primarily on grammar can make Sanskrit tough and tedious for new learners.
In an effort to make Sanskrit learning exciting and more accessible to wider audience, we developed an Automatic Speech Recognition (ASR) model for Sanskrit and this paper discusses the approach that was employed in building such a model. The developed ASR model could be used for transcription, practising pronunciation, etc. 

While designing the model we noted the inherent linguistic complexities of Sanskrit that challenge ASR. For instance:
(\lowerromannumeral{1})~Unlike languages with fixed word forms, Sanskrit allows for the dynamic combination of morphemes, resulting in multiple pronunciations for a single concept~\cite{malik1995}. This variability, combined with the lack of standardised pronunciation rules, poses difficulties for ASR models in accurately mapping sounds to words~\cite{rao2014}. 
(\lowerromannumeral{2})~Sanskrit's diverse phonetic inventory possesses a rich set of sounds, including retro-flexes and breathy consonants, which are absent in many modern languages~\cite{rao2014}. Accurately distinguishing these subtle nuances, especially in noisy environments, is a challenge for ASR models trained in simpler languages~\cite{sharma2018}. (\lowerromannumeral{3})~Sanskrit pronunciation has undergone significant changes over time, creating further discrepancies between written and spoken forms\cite{witzel2003}. This necessitates ASR models to be capable of handling all these nuances and complexities ingrained in the linguistics of the language.

Traditional statistical approaches for ASR such as Acoustic Models and Language Models often fall short while working on Sanskrit language for the reasons mentioned above. In this paper, we employ transformer architecture~\cite{vaswani2017attention} for the task of ASR for Sanskrit. It is essentially a sequence-to-sequence model that is capable of addressing to a large extent the above-mentioned challenges posed by the Sanskrit language for building an ASR. It utilises a self-attention mechanism that analyses relationships between various sounds within a speech segment rather than working on isolated sounds, leading to more accurate recognition of subtle variations like retroflex vs. dental consonants. 


The rest of the paper is organised as follows: Section~\ref{sec:lit} reviews the related work available in the literature for the task at hand. Section~\ref{sec:pm} details our proposed methodology.  The results of our experiments are presented in Section~\ref{sec:res}.  Finally, Section~\ref{sec:con} summarises our conclusions and explores potential avenues for future work.

\section{Literature Review} \label{sec:lit}
Early research in Sanskrit ASR primarily employed statistical models like Hidden Markov Models and Gaussian Mixture Models  \cite{hmmasr}\cite{pokhariya2014sanskrit}. These models managed to map basic acoustic patterns in controlled environments, such as recitations of the Ramayana. However, the complexities of real-world scenarios – background noise, diverse speaker accents, and spontaneous speech – proved insurmountable. These models struggled amidst the rich linguistic intricacies of spoken Sanskrit. They were unable to effectively capture the language's sophisticated grammar rules and vast vocabulary.

Statistical-based approaches for ASR such as Acoustic Models (AMs) and Language Models (LMs) have been used in several works. For instance, in~\cite{adiga2021automatic} the authors propose a linguistically inspired new syllable-level unit segmentation technique, which they term as \textit{vowel segmentation}.  Three encoding schemes for tokenization are investigated: word-based encoding, Byte Pair Encoding (BPE), and vowel segmentation. The best reported Word Error Rate (WER) is 21.94\% using the Sanskrit Library Phonetic basic encoding scheme (SLP1) script with BPE encoding for the LM unit and Grapheme encoding for the AM unit on a dataset that they specially created for Sanskrit ASR evaluation called Vāksañcayaḥ. Nonetheless, these models may fail to capture contextual relationships when processing on long speech audios spanning several seconds.

Many researchers have employed deep learning techniques, which offer advanced analytical tools for the task of ASR. For instance, in~\cite{ctcbasedasr}, the authors train a neural network using the Connectionist Temporal Classification (CTC) approach for the task of automatic speech recognition for Sanskrit. 
Notably, the model uses a single bidirectional Long Short-Term Memory (LSTM) layer for direct character sequence prediction, bypassing separate phoneme classification stages leading to improved efficiency and accuracy.
To address the issue of limited training data, the research employs spectrogram augmentation. 
The model achieves a word error rate (WER) of 7.6\% on their proprietary dataset.

Another study~\cite{g2p} evaluates four grapheme-to-phoneme (G2P) conversion schemes for ASR for Sanskrit using a proprietary speech corpus of about 15 hours. 
They use a factorised time-delay neural
network trained on an objective function using a modified version of SLP1 script to solve the ASR for Sanskrit achieving a WER of 8.4\%.

The work of~\cite{speech_recognition_2023} evaluates syllable-based modelling units for end-to-end speech recognition in Indian languages, utilising various text representations such as native script, SLP1 encoding, and syllables. The results on the Vāksañcayaḥ dataset indicate that syllable-based BPE units perform effectively in monolingual speech recognition but not in cross-lingual transfer learning, where SLP1 character units demonstrate superior performance. According to their syllable-based modelling units, different languages have different syllable-based representations so using one syllable-based representation in one language may not be able to represent the syllable of other languages. Their results also show that the Sanskrit-based syllable representation does not perform well in Tamil and Kannada languages. Nonetheless, their paper reveals that syllable-BPE representation achieves the best results for the Sanskrit dataset. The limitation is that the Syllable-based tokens do not support cross-lingual transfer learning due to varying syllable distributions and associations across languages.

In recent years, transformers have emerged as a foundational architecture across various fields. Their capacity to model long-range dependencies and capture intricate relationships has driven significant advancements in natural language processing and beyond. Leveraging their success, this paper presents the development of a Sanskrit ASR model using the Transformer architecture. Given the limited availability of Sanskrit data, a transfer learning approach was employed. Hyperparameter tuning substantially enhanced the model's performance.
The details of the proposed methodology are described in the next section.

\section{Proposed Methodology} \label{sec:pm}

Our model is based on OpenAI's Whisper, a state-of-the-art pre-trained transformer architecture for Automatic Speech Recognition (ASR). To perform ASR for Sanskrit, we employ a transfer learning approach. 
Specifically, we trained the Whisper model for ASR on a new langauge, Sanskrit,  utilising the Vāksañcayaḥ dataset~\cite{adiga2021automatic}, which contains speech audio files rendered exclusively in Sanskrit. Such audio inputs were not encountered by the pre-trained Whisper model previously.
We leveraged on the existing linguistic knowledge that Whisper had learnt after being trained on a huge English audio corpus while adapting it to the specificities of Sanskrit grammar, pronunciation and prosody through our training.  This transfer learning approach was crucial due to the scarcity of data available for training Sanskrit ASR models.

\begin{figure}[ht]
	\centering
	\includegraphics[width=0.45\linewidth]{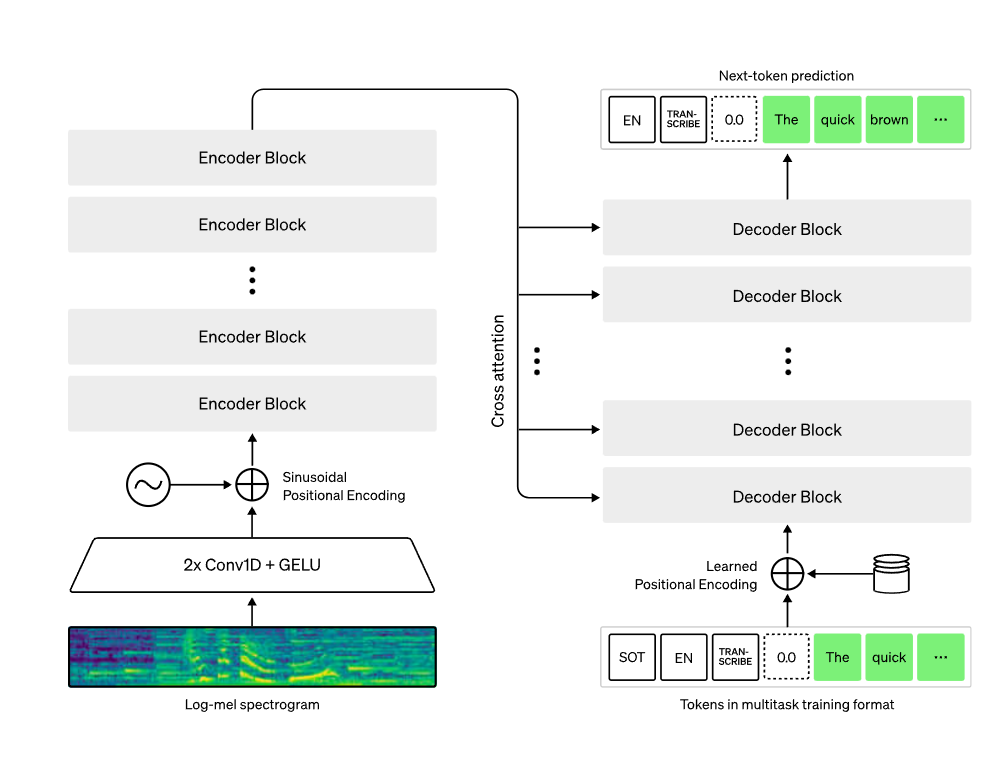}
	\caption{Whisper Model Architecture~\cite{radford2022robust}}
	\label{fig:whisper_model}
\end{figure}

Before delving into the implementation details, we present a brief overview of the Whisper model~\cite{radford2022robust}. 
The model marks a significant achievement in ASR technology. Trained on a massive dataset of labeled audio and text exceeding 680,000 hours, Whisper demonstrates exceptional accuracy and robustness in transcribing spoken language. Although, the model is capable of several other speech processing tasks such as multilingual speech translation, language identification, voice activity detection, we primarily used it for the purpose of ASR for Sanskrit. The model architecture of Whisper family is shown in Fig.~\ref{fig:whisper_model}

%

\begin{table}
	\centering
	\begin{tabular}{|c|c|c|c|c|}
		\hline
		\textbf{Model} &\textbf{Layers} & \textbf{Width} & \textbf{Heads} & \textbf{Parameters (M)} \\
		\hline
		Tiny  & 4  & 384 & 6 & 39  \\
        \hline
        Base  & 6  & 512 & 8& 74  \\
        \hline
        Small & 12 & 768 &12 & 244 \\
        \hline
        Medium & 24 & 1024 & 16 & 769 \\
        \hline
        Large & 32 & 1280 & 20  & 1550 \\
		\hline 
	\end{tabular}
\vspace{2ex}
 \caption{Key architecture details of the Whisper model family\cite{radford2022robust}}
	\label{tab:whisper-models}
\end{table}

Table~\ref{tab:whisper-models} outlines key architectural details of various transformer models under Whisper family. The number of encoder and decoder layers is specified by the ``Layers" column,
while the dimensionality of the hidden state, post-token embedding, is indicated by ``Width". The ``Heads" column denotes the count of attention heads per layer, enabling the model to focus on different parts of the processed spectrogram sequence. The final column, ``Parameters (M)," denotes the total number of trainable parameters in the model, serving as a measure of its complexity.

We employed a transfer learning approach to train the Whisper model for Sanskrit speech recognition. Specifically, we fine-tuned both the Whisper Small and Medium models (see Table~\ref{tab:whisper-models}) on the Vāksañcayaḥ Sanskrit dataset~\cite{adiga2021automatic}. We call the resulting models as Whisper Transfer Learned (WTL) models, which were tailored to capture linguistic features,
patterns, and representations of Sanskrit speech.

\subsection{Train Phase}

During training,
the model receives both speech audio and its corresponding Devanagari script as inputs. The audio undergoes preprocessing: resampling, normalization, and windowing, followed by conversion into a log Mel spectrogram. This spectrogram is processed by convolutional neural networks and combined with sinusoidal position embeddings before being fed to the encoder. 
The encoder processes the input sequence and generates hidden states that capture relevant information from the speech audio. These hidden states are then passed to the decoder as indicated in Figure~\ref{fig:whisper_model}.

The other input, Devanagari transcript (ground truth), is preprocessed before being sent to the decoder.
It is tokenized, word embedded, and position encoded. This allows the decoder to iteratively generate tokens and refine its predictions based on the encoder's output and previously generated text. This process continues until the decoder generates an end-of-sequence token, signifying the transcription's completion.

The model parameters are optimised to minimise the transcription error rate on a validation set, assessed using the WER metric.

\subsection{Test Phase}

During testing,
the speech audio is the sole input. The encoder processes it as in training. The decoder, however, starting with a start-of-sequence token, iteratively extends the sequence, token by token, based on the encoder's output and previously generated text to produce the transcribed text. For more details on the transformer architecture, please refer to~\cite{vaswani2017attention}.
We built our model using Hugging Face, an open-source platform, that provides a streamlined framework for utilising the Whisper model and its associated components~\cite{huggingface}.

\subsection{Hyper-parameter Optimisation}

As with any other neural network training, choosing the right set of hyper-parameters is crucial for the optimal performance of the model. For our model, we employed a {randomised search algorithm} on the key hyper-parameters used in the model. The range of these hyper-parameters that was given to the search algorithm are as follows:

\begin{enumerate}
    \item \textit{MLP Dropout:} Range was set from 0.2 to 0.6 to control the dropout rate within the model's multi-layer perceptrons (MLPs) to prevent over-fitting.
    \item \textit{Attention Dropout:} Range was set from 0.2 to 0.6 to regulate the dropout rate within the model's attention mechanism for optimal information flow and to prevent over-fitting.

    \item \textit{Learning Rate:} Fine-tuning was based on the model size, with ranges of 1e-5 to 3e-5 for the WTL-Small model and 5e-5 to 8e-5 for the WTL-Medium model as recommended in~\cite{radford2022robust}.

    \item \textit{Optimiser:} Various optimisers were evaluated, including \textit{adagrad, adamw, adam\_hf,} and \textit{adam}, to identify the most effective algorithm for updating the model parameters.

    \end{enumerate}
    
    We also explored various learning rate schedulers including \textit{linear}, \textit{linear-warmup}, \textit{cosine}, and \textit{reduce-on-plateau} to dynamically adjust the learning rate for optimal convergence during the training.

The optimal configuration of the hyper-parameters within these ranges were identified for both the WTL-small and WTL-medium models. This hyper-parameter optimisation on the Vāksañcayaḥ dataset \cite{adiga2021automatic} along with the transfer learning employed on the pre-trained Whisper models were effective in capturing the characteristics of the Sanskrit language and contributed to superior performance than those obtained with  baseline configurations (ie. using the default configuration, with no hyper-parameter tuning).

For optimal performance and analysis, a suite of additional libraries was utilised from Hugging Face. For instance,  the Accelerate library from Hugging Face was leveraged for efficient and speedy training on GPUs, benefiting from its optimised training mechanisms. The Weights \& Biases library was used to track each training run with its corresponding hyper-parameter settings, allowing for efficient analysis and identification of the best-performing configurations for both WTL-small and WTL-medium models.

\begin{figure}
  \centering
  \includegraphics[width=0.40\linewidth]{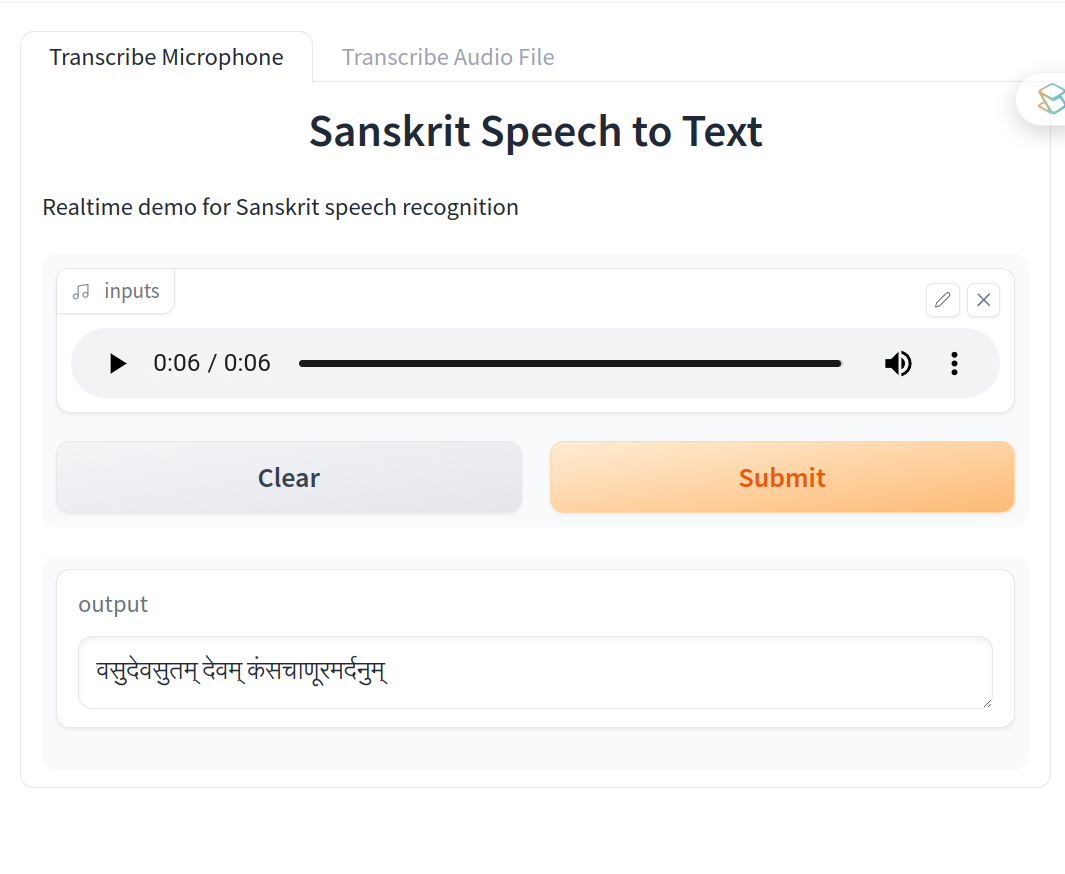}
  \caption{A screenshot of the online demo built using Gradio library.}
  \label{fig:demo}
\end{figure}

To showcase the accuracy and robustness of our transfer-learned Whisper model for Sanskrit ASR, we developed an online demo using the Gradio library \cite{gradio_docs} on the Hugging Face Spaces. A screenshot of the online demo is shown in Figure~\ref{fig:demo}. This interactive platform allows users to evaluate the model's capabilities first-hand, where the spoken Sanskrit speech audio is transcribed with ease.

The demo further expands its accessibility by supporting both microphone recordings and audio file uploads. This caters to diverse preferences and application scenarios, ensuring the model's usability for a wider audience. Additionally, the ability to automatically split longer audio files exceeding 30 seconds \cite{radford2022robust} into manageable chunks demonstrates the model's proficiency in handling extended speech while maintaining consistent performance. To access the Sanskrit ASR demo, visit the page\footnote{ {\footnotesize \url{https://cs.rkmvu.ac.in/research/demos/}}}.

\section{Results} \label{sec:res}

To evaluate our WTL model we used a publicly available dataset for Sanskrit ASR called Vāksañcayaḥ~\cite{adiga2021automatic}. It has over 78 hours of audio recordings, comprising of 45,953 sentences spoken by 27 speakers with diverse linguistic backgrounds. This variety ensures the corpus captures the natural variations encountered in spoken Sanskrit. The dataset includes transcripts in both Devanagari script and SLP1 format.
The texts are sourced from sacred verses, contemporary narratives, radio programs, and extemporaneous speeches, providing a well-rounded representation of spoken Sanskrit across various contexts. Furthermore, it has a separate out-of-domain test set featuring speakers included neither in the training nor validation dataset which serves as a crucial benchmark for assessing the generalisability of ASR models trained on Vāksañcayaḥ. We used the training, validation and testing split as recommended by the authors of the dataset.

We report our results on the small and medium sized models (see Table~\ref{tab:whisper-models}). As mentioned earlier, hyper-parameter optimisation was crucial to improve the performance of the models for this low-resource language.  We used the WER as our metric in our evaluations. The results obtained from our experiments are shown in Table~\ref{tab:res}.  We note that the WERs of both these models on the test set are significantly lower than that of the model proposed in~\cite{adiga2021automatic} thereby demonstrating the efficacy of our transfer learning approach employed on this low resource language. 
In particular, we note a percentage reduction of 29.8\% in WER by using the WTL-Medium model when compared to that of the model proposed in~\cite{adiga2021automatic}. Also,
the WTL-Medium  model exhibited a robust performance on the out-of-domain test set, with a WER of 37\% while the model proposed in~\cite{adiga2021automatic} obtained 44\%, achieving a percentage reduction of 15.7\% in WER. The result demonstrates the proposed model's resilience against noise and unfamiliar speech patterns. Such a robust performance indicates the model's potential for real-world applications beyond controlled test environments.

\vspace{1.2ex}

\begin{table}[htbp]
\centering
\begin{tabular}{|c|c|c|c|}
\hline
\textbf{Model}&\multicolumn{2}{|c|}{\textbf{Word Error Rate}} \\
\cline{2-3} 
\textbf{} & \textbf{\textit{Test(\%)}}& \textbf{\textit{OOV$^{\ast}$(\%)}} \\
\hline
WTL-Small& 19.85 & 44.85  \\
\hline
 WTL-Medium & 15.42 & 37.22 \\
\hline
Method proposed in \cite{adiga2021automatic} & 21.99 & 44.16 \\
\hline
\multicolumn{3}{l}{$^{\ast}$ Out of Domain Set}
\end{tabular}
\vspace{2ex}
\caption{Word Error Rates (WER) obtained on the Vāksañcayaḥ test dataset~\cite{adiga2021automatic}}
\label{tab:res}
\end{table}

It is noteworthy that while working on the small speech audio dataset, Vāksañcayaḥ, we encountered the common phenomenon of model overfitting. To address this issue, we adopted the following measures:

\begin{itemize}
    \item \textit{Early Stopping:} We monitored the validation loss and stopped training when it started to increase; 
    \item \textit{Regularisation:} We incorporated techniques such as dropout and weight decay to control model complexity
    \item \textit{Data Augmentation:} We employed data augmentation techniques. A subset of audio files was randomly selected and subjected to transformations such as pitch shifting, time stretching, and the addition of silence or background noise to enhance data diversity.
\end{itemize}

We trained the WTL-Medium model on a single NVIDIA Quadro GV100 GPU, and the WTL-Small model on a single NVIDIA GeForce RTX 2080 Ti GPU. Average training times were approximately 36 and 14 hours,
respectively. The system ran on Linux kernel version 5.15.0-86-generic-x86\_64 with glibc2.31 and Python 3.11.5. To conveniently track the hyper-parameters and to visualise various plots in our experiments, we utilised Weights \& Biases CLI version 0.15.11 platform~\cite{wandb}.

\section{Conclusions and Future Work} \label{sec:con}
This paper employed a transfer learning approach to train a  Whisper model, which is pre-trained on a massive English-specific corpus, for Sanskrit ASR using the Vāksañcayaḥ dataset. This leveraged the model's inherent linguistic knowledge (learnt due to the previous training) while adapting it to the specificities of Sanskrit grammar, pronunciation and prosody (learnt from the training on the Vāksañcayaḥ dataset). 
Furthermore, optimising the model's hyper-parameters led to significant improvements in its performance.

Our best model achieved a WER of 15.42\% on the Vāksañcayaḥ test dataset, surpassing previous benchmarks. Additionally, it demonstrated robust performance against noise and various speech accents that existed in the out-of-domain test set.

With better transcription accuracy compared to the previously existing models, our model can be used to transcribe historical recordings and literary works, thereby improving accessibility for researchers, educators, and enthusiasts. This capability to analyse large audio datasets can drive research in language evolution, cultural studies, and other fields enriched by Sanskrit.

As part of the future work, the following avenues may be investigated:

\begin{itemize}

    \item Paucity of data: Transformer models require vast amounts of high-quality data for optimal performance. Thus, the creation of new Sanskrit datasets with vast and diverse speech content is vital. We noted that the dataset we employed was limited in size especially for transformer based models.
    \item Sanskrit-Specific Tokenization: Our current model utilises Whisper's default multilingual tokenization. Developing a customised tokenizer specifically designed for Sanskrit's unique morphology and phonology could significantly improve accuracy. 
    \item Better evaluation metrics:  WER may not fully capture the nuances of Sanskrit, where homonyms and words with variant spellings (\textit{sandhi}) convey the same meaning. Exploring alternative metrics which incorporate semantic knowledge of Sanskrit could provide a more accurate assessment of the model's performance.
    \item Multimodal Learning: Integrating additional modalities, such as lip-reading or visual context, into the model could potentially improve its performance, particularly in noisy or ambiguous situations. 
    \item Domain-Specific Adaptation: Fine-tuning the model for specific domains, such as \textit{Vedas}, contemporary dialogues, or technical lectures, could further improve accuracy and cater to specialised research needs. 
\end{itemize}

By working on these possible future avenues, we can further improve the performance and robustness of our model.

\vspace{5ex}

\bibliography{reference.bib}{}
\bibliographystyle{plain}

\end{document}